\documentclass[preprint,12pt]{elsarticle}
\usepackage{etex}
\usepackage{amssymb}
\usepackage{amsmath}
\usepackage{booktabs}
\usepackage{pifont}
\usepackage{makecell}
\usepackage{paralist}
\usepackage{hyperref}

\journal{Neurocomputing}

\begin{document}

\begin{frontmatter}



\title{Milmer: a Framework for Multiple Instance Learning based Multimodal Emotion Recognition}


\author[label1]{Zaitian Wang}
\author[label1]{Jian He}
\author[label1]{Yu Liang\corref{cor1}\thanks{Corresponding author}}
\ead{yuliang@bjut.edu.cn}
\author[label1]{Xiyuan Hu}
\author[label2]{Tianhao Peng}
\author[label1]{Kaixin Wang}
\author[label3]{Jiakai Wang}
\author[label1]{Chenlong Zhang}
\author[label1]{Weili Zhang}
\author[label1]{Shuang Niu}
\author[label1]{Xiaoyang Xie}

\cortext[cor1]{Corresponding author}

\affiliation[label1]{
    organization={Beijing University of Technology},
    city={Beijing},
    postcode={100124},
    country={China}
}
\affiliation[label2]{
    organization={Beihang University},
    city={Beijing},
    postcode={100191},
    country={China}
}
\affiliation[label3]{
    organization={Zhongguancun Laboratory},
    city={Beijing},
    postcode={100194},
    country={China}
}

\begin{abstract}
Emotions play a crucial role in human behavior and decision-making, making emotion recognition a key area of interest in human-computer interaction (HCI). This study addresses the challenges of emotion recognition by integrating facial expression analysis with electroencephalogram (EEG) signals, introducing a novel multimodal framework-Milmer. The proposed framework employs a transformer-based fusion approach to effectively integrate visual and physiological modalities. It consists of an EEG preprocessing module, a facial feature extraction and balancing module, and a cross-modal fusion module. To enhance visual feature extraction, we fine-tune a pre-trained Swin Transformer on emotion-related datasets. Additionally, a cross-attention mechanism is introduced to balance token representation across modalities, ensuring effective feature integration. A key innovation of this work is the adoption of a multiple instance learning (MIL) approach, which extracts meaningful information from multiple facial expression images over time, capturing critical temporal dynamics often overlooked in previous studies. Extensive experiments conducted on the DEAP dataset demonstrate the superiority of the proposed framework, achieving a classification accuracy of 96.72\% in the four-class emotion recognition task. Ablation studies further validate the contributions of each module, highlighting the significance of advanced feature extraction and fusion strategies in enhancing emotion recognition performance. Our code are available at https://github.com/liangyubuaa/Milmer.
\end{abstract}

%


\end{frontmatter}

\section{Introduction\label{sec:section1}}
Emotions, as an integral part of human daily life, significantly influence individual behavior, decision making, social interactions, and mental health. In recent years, with the rapid advancement of human-computer interaction (HCI) technologies, emotion recognition has emerged as a highly prominent research area. Extensive studies have explored this topic through various data modalities. The development of deep learning methods, coupled with the utilization of large-scale multimodal datasets such as DEAP \cite{deap} have revolutionized the field of emotion recognition. These advancements have paved the way for detecting and interpreting human emotions with unprecedented accuracy.

Emotion recognition relies broadly on two primary modalities of human data: visual data such as images or videos, and physiological signals such as electroencephalogram (EEG). Research in this field has first focused on vision-based approaches, using facial images and videos, as evident in prior studies \cite{huangEnhance,FER2013,nie2020c,husformer}. Facial expressions, as direct and easily captured indicators of human emotions, have been extensively studied. However, facial expressions in datasets commonly used in facial emotion recognition often exhibit exaggerated or highly recognizable features. These expressions, while easily identifiable, are not representative of real-life emotional expressions, where emotions are not always expressed in an exaggerated or obvious way. In fact, people often display little to no facial expression in response to their emotions. This discrepancy poses a significant challenge in accurately recognizing emotions in real-world scenarios and highlights a major weakness in the field of facial emotion recognition.

In contrast, physiological signals such as EEG offer distinctive insights into human emotions by reflecting intrinsic responses that occur instinctively and are not subject to conscious control, providing a more accurate representation of genuine emotions \cite{liang2024fetcheeg,shu2018review,zcl}. Despite their potential, the adoption of deep learning strategies for physiological signals remains limited, hindered by challenges such as hardware inconsistencies, the differences in data preprocessing, and the high cost of bio-sensing data acquisition.

Beyond single modality approaches, recent studies have highlighted the advantages of multimodal affective computing techniques, which integrate vision and physiological data to achieve improved emotion recognition performance \cite{zhang2020emotion,startingArticle,greatHelp}. Each modality contributes unique strengths that complement the other, and their fusion enhances the quality of the extracted features, resulting in a more comprehensive representation of human emotions. Each modality contributes unique strengths that complement each other, and their fusion enhances the quality of the extracted features, resulting in a more comprehensive representation of human emotions, which significantly improves the accuracy compared to single modality approaches\cite{pan2023review}. Moreover, in multimodal datasets, facial expression data are typically derived from videos of participants watching stimulus videos, where their expressions are more subdued and natural. This makes the facial expressions in multimodal studies closer to real-life scenarios compared to those captured in single modality datasets, where expressions are often more exaggerated. 

However, despite the proven value of using EEG and facial expressions in multimodal emotion recognition, there remains a substantial gap in research in this area, leaving considerable opportunities for further exploration\cite{pan2023review}. Current research tends to focus more on feature extraction within each modality independently, while neglecting the crucial task of modality fusion. Many approaches rely on simple methods such as concatenation or decision-tree-based fusion techniques, yet EEG and facial expressions represent two fundamentally different modalities. Such simplistic fusion strategies often lead to the loss of valuable information and fail to fully utilize the complementary features extracted from both modalities. This study argues that these limitations are a significant reason why existing models have not fully realized their potential, and that more sophisticated, comprehensive fusion methods should become the standard in future research.

It is worth noting that, due to the inherently higher complexity of visual information compared to EEG signals, most prior research has placed a stronger emphasis on facial expression analysis. However, this study argues that facial expressions in multimodal datasets tend to be more constrained and neutral, closely resembling real-world conditions where distinguishing subtle emotions becomes particularly challenging. In such scenarios, EEG signals play an irreplaceable role, as they can capture underlying emotional states that are not easily reflected in facial expressions. This highlights the equal importance of both modalities and underscores the critical need for effective multimodal fusion strategies\cite{vilt}. Unfortunately, previous studies often overlook this aspect, applying overly simplistic fusion techniques that fail to effectively combine these two fundamentally different types of data.

In the feature extraction process for the facial expression modality, CNN-based architectures are predominantly used in this field, often capturing localized features rather than global ones. Additionally, research frequently relies on single facial expression images rather than multiple images over time to assess emotions, thereby failing to fully utilize the available information. These two limitations reduce the quality of visual feature representation and, ultimately, impact emotion classification performance.

To conclude, several limitations have been identified in the current field: (1) Despite its potential, the field remains underexplored, with relatively few studies dedicated to fully integrating facial expression and EEG signals for emotion recognition. (2) Most research efforts focus primarily on feature extraction from individual modalities, with limited investigation into effective multimodal fusion strategies. (3) Current approaches to facial expression feature extraction rely on outdated methods and fail to consider using multiple images over time for emotion recognition, missing important temporal information.

To address these challenges, this paper proposes a hybrid framework named Milmer for \underline{M}ultiple \underline{I}nstance \underline{L}earning
based \underline{M}ultimodal \underline{E}motion \underline{R}ecognition, which includes an EEG preprocessing module, a facial feature extraction and balancing module, and a transformer-based modality fusion module. The key contributions of this paper are as follows:

\begin{enumerate}
\item Multimodal emotion recognition using facial expressions and EEG signals remains underexplored. This study fills this gap by providing a comprehensive exploration of facial expression and EEG signal integration, along with open-sourcing the code to facilitate further research in this area;
\item This paper proposes a transformer-based modality fusion approach, which more effectively integrates the visual and physiological modalities, overcoming the limitations of previous, simplistic fusion methods;
\item This study employs a multi-instance learning approach to extract useful information from multiple facial expression images over time. It also leverages a pre-trained Swin Transformer, fine-tuned on emotion-related datasets, to extract high-quality features\cite{swin}. Additionally, a cross-attention mechanism is used to balance the features of facial images, ensuring a more comprehensive and effective fusion of both modalities.
\end{enumerate}

The rest of this paper is organized as follows. Section~\ref{sec:section2} introduces related work on the topic. Section~\ref{sec:section3} outlines the structure of the proposed method. Section~\ref{sec:section4} details the experimental setup and discusses the results. Finally, Section~\ref{sec:section5} draws the conclusions and presents future research directions.

\section{Related Work\label{sec:section2}}

\subsection{EEG-Based Emotion Recognition}
Electroencephalogram (EEG)-based emotion recognition utilizes brain activity signals to classify emotional states, offering an intrinsic and unconscious reflection of emotional responses. Deep learning models such as convolutional neural networks (CNNs) \cite{schirrmeister2017deep} and recurrent neural networks (RNNs) \cite{zheng2017identifying} have been extensively applied for feature extraction and classification. More recently, transformer-based approaches \cite{wang2022eegtransformers,liang2024fetcheeg} have demonstrated strong potential in modeling temporal dynamics and long-range dependencies in EEG signals. Advanced frameworks such as graph neural networks (GNNs)\cite{Gan2020eeg} have also been proposed to capture the spatial interrelations among EEG channels effectively.

However, EEG recording is associated with several challenges. Key obstacles include the presence of internal and external artifacts, such as eye movements and muscle activity, which can interfere with signal accuracy. Furthermore, recording EEG data for extended periods can introduce noise and variability, which complicates the extraction of consistent emotional features\cite{eegReview}. These factors limit the effectiveness of EEG as a standalone modality for emotion recognition, highlighting the need for complementary approaches.

\subsection{Facial Emotion Recognition}
Facial emotion recognition (FER) involves analyzing human facial expressions from images or videos to classify emotions. Deep learning methods, particularly convolutional neural networks (CNNs), have significantly advanced this field by automating facial feature extraction. For static images, 2D CNNs effectively extract spatial features, achieving high accuracy in emotion classification \cite{breuer2017deep}. For video data, methods such as Conv3D \cite{fan2016video} and ConvLSTM \cite{huang2018end} are employed to capture both spatial and temporal information, improving recognition performance. Recently, transformer architectures \cite{zhao2021former,huang2021facial} have gained attention for their ability to encode context-aware features through attention mechanisms, offering superior results in dynamic emotion analysis.

These advancements have made FER a cornerstone of emotion recognition research. However, relying solely on facial images or videos has inherent limitations, particularly in real-world scenarios where emotions are often subtle, neutral, or influenced by contextual factors that are not captured visually \cite{facialReview}. The absence of complementary data, such as physiological signals, restricts FER methods from fully understanding the underlying emotional states. These limitations highlight the necessity of exploring multimodal approaches, which integrate diverse data sources to achieve more accurate and robust emotion recognition.

\subsection{Multimodal Emotion Recognition}
Multimodal emotion recognition, which integrates facial expressions and EEG signals, has gained significant attention due to its ability to provide more accurate emotion classification by combining complementary information. However, research in this area still lacks sufficient exploration of effective fusion techniques, which limits the potential of multimodal models.

Several approaches have been proposed to fuse EEG and facial expression data. Early methods often relied on simple fusion strategies such as decision trees or voting mechanisms to combine the outputs of each modality \cite{tan2021multimodal,startingArticle}. While these approaches are straightforward, they fail to effectively utilize the rich, modality-specific features extracted from each source, often resulting in suboptimal performance. Methods like Deep Canonical Correlation Analysis (DeepCCA), have attempted to improve feature correlation between modalities, yet they still fall short of fully exploiting the complex, individual features of each modality \cite{muhammad2023bimodal}. Low-rank fusion techniques have also been explored, offering a more efficient way to combine modality-specific features \cite{liu2018efficient}. While these methods are computationally efficient, they are not advanced enough to capture the complex interactions between the two modalities.

Concatenation-based methods, which combine the features of both modalities into a single vector, are also widely used but can lead to information loss \cite{jung2019utilizing,greatHelp,huangEnhance}. In contrast, more sophisticated methods, such as cross-attention mechanisms in transformers, allow the model to focus on the most relevant features from each modality, enhancing fusion accuracy and improving overall performance \cite{husformer}. Despite these advancements, the field still lacks comprehensive, effective fusion strategies that can fully harness the strengths of both EEG and facial expression data, highlighting the need for further research in this area.

\subsection{Multiple Instance Learning}
Multiple Instance Learning (MIL) is a variation of supervised learning in which a class label is assigned to a bag of instances rather than individual instances. Unlike traditional supervised learning, where each input (such as an image) is labeled with a specific category, MIL is applied in scenarios where the labeling is more ambiguous. In MIL, a "bag" contains multiple instances, and only the overall class of the bag is provided, not individual labels for each instance. This approach is particularly useful when data annotations are weak or incomplete, a common occurrence in real-world tasks.

Traditional MIL typically relies on pooling methods like max pooling or average pooling to combine the individual instances in a bag and make a prediction. However, these pooling techniques often fail to capture the importance of specific instances, as they treat all instances equally. In contrast, attention-based MIL (AMIL) \cite{amil}introduces an attention mechanism that assigns different weights to instances, allowing the model to focus more on the most relevant instances while ignoring less informative ones. This approach significantly improves MIL performance by enabling more precise feature extraction and decision-making.

In the field of multimodal emotion recognition using EEG and facial expressions, the typical approach involves associating a short segment of EEG data (e.g., 3 seconds) with a single facial expression image, which is then used to represent the entire 3-second window. However, this approach is inherently flawed as it may overlook valuable temporal and subtle details present within the sequence of images. A 3-second video, for instance, should be understood as a "bag" of instances, with each individual frame or image representing a potential instance. Relying on just one image to represent the entire bag could result in the loss of critical information, especially in capturing emotional variations that occur over time.

By adopting MIL, we can leverage all instances within the video to create a more balanced and accurately represent the entire segment. MIL allows for a more comprehensive understanding by treating the entire video as a bag of instances, thus enhancing the emotion recognition process. Despite the potential benefits, very few studies have recognized the value of MIL in this context \cite{rao2016multi,romeo2019multiple}. This research aims to incorporate MIL to better capture and represent the emotional content in the facial expression modality.

\section{Methods\label{sec:section3}}

\begin{figure}[!t]
\centering
\includegraphics[width=\linewidth]{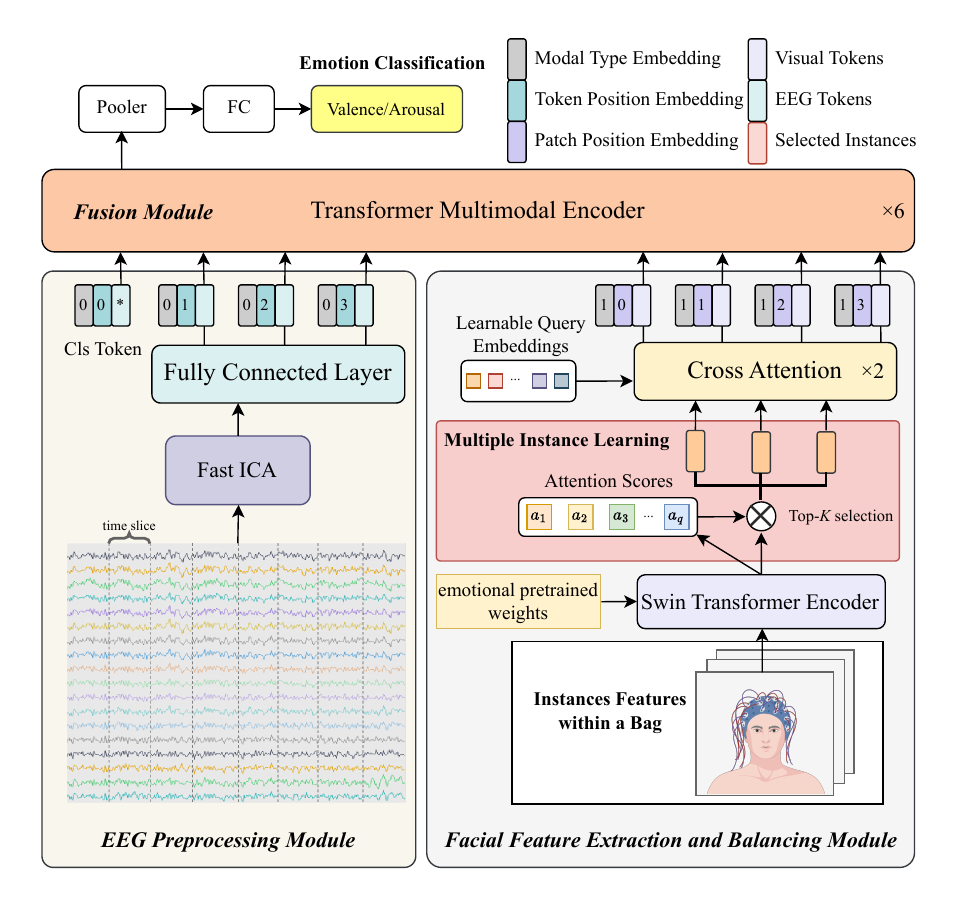}
\caption{The overview of our framework, consisting of a EEG preprocessing module, a facial feature extraction and balancing module, and a modality fusion module.}\label{fig1}
\end{figure}

\subsection{Overview\label{sec:section31}}
As discussed earlier, many existing methods overlook the importance of effective modality fusion, relying on simple concatenation or decision-tree-based approaches that fail to fully exploit the complementary nature of different modalities. Transformer\cite{attention}, an emerging neural network architecture initially developed for machine translation tasks, has recently achieved remarkable success in the field of natural language processing (NLP). However, its potential application to emotion recognition tasks involving EEG and facial modalities remains underexplored.

In this work, we propose a transformer-based framework to achieve effective cross-modal fusion. As illustrated in Fig.~\ref{fig1}, the proposed architecture consists of four main modules: 
\begin{inparaenum}[(1)]
    \item EEG preprocessing module,
    \item facial feature extraction and balancing module, and
    \item modality fusion module.
\end{inparaenum}

In the first module, noise and artifacts in the EEG signals are filtered out to ensure clean input data. In the second module, Swin Transformer\cite{swin} is used to extract features from facial images. A MIL approach is then introduced to better represent the facial expressions by selecting the most representative frames. To further reduce the feature dimension and highlight key information, a cross-attention mechanism\cite{attention,blip2} is applied to the Swin Transformer output. Then, both the EEG and facial features are embedded and fed into a transformer for cross-modal fusion. Finally, the fused features, which encapsulate frequency, temporal, and spatial information, are passed through a fully connected classifier to produce the final emotion classification results. The details of these three modules are explained in the subsequent sections.

\subsection{EEG Preprocessing Module\label{sec:section32}}
The majority of EEG signals are concentrated within the frequency range of 1 Hz–50 Hz. Therefore, a bandpass filter with a passband of 1 Hz–50 Hz is applied in the proposed model. This filtering step serves two primary purposes: \begin{inparaenum}[(1)]
    \item removing low-frequency baseline drift, electrocardiographic (ECG) interference, and other high-frequency noise, and 
    \item effectively mitigating the most prominent power line interference, which typically occurs at 50 Hz in China.
\end{inparaenum}

However, EEG signals often overlap with electrooculogram (EOG) and electromyogram (EMG) signals within the same frequency band, making a single bandpass filter insufficient for completely eliminating these artifacts. To address this issue, Independent Component Analysis (ICA) is employed. ICA is a computational technique that decomposes a multivariate signal into a set of statistically independent components. By maximizing the non-Gaussianity of the sources, ICA separates mixed signals into independent sources, which helps to isolate and remove artifacts such as EOG and EMG from EEG signals.

The ICA process involves estimating a matrix that unmixes the data into independent components, which are then identified and removed based on their spatial and temporal characteristics. Specifically, the Fast ICA algorithm from the MNE library\cite{mne} is utilized to perform this decomposition. For ICA processing, continuous EEG data from a single subject (without segmenting it into epochs) is used, with all other settings left at their default values in the MNE library. By combining bandpass filtering and ICA, this preprocessing module effectively ensures cleaner EEG signals for subsequent tasks.

\subsection{Facial Feature Extraction and Balancing Module\label{sec:section33}}

This study significantly advances the visual feature extraction process compared to prior research in the field. First, in contrast to traditional CNN-based methods prevalent in the field, we utilize the advanced Swin Transformer fine-tuned on emotion classification dataset as the backbone for feature extraction. 

Previous studies typically align a single facial image with a segment of EEG data (commonly 3 seconds), assuming that a single frame can represent the entire video segment. However, it is difficult to determine which frame best encapsulates the information from a 3-second video. To address this limitation, we introduce MIL into this domain. By extracting multiple frames from the video segment and leveraging MIL, the most representative K frames are selected. This approach ensures a more comprehensive and nuanced representation of the facial modality, capturing richer emotional information.

Lastly, this study addresses the critical challenge of balancing the contributions of the modality during the fusion stage. Selecting multiple frames through MIL inevitably introduces far more visual tokens compared to the EEG modality, which could lead to the model disproportionately focusing on facial features while neglecting EEG information. To mitigate this imbalance, we employ a cross-attention-based mechanism, enabling the K frames to interact and learn from each other while reducing their dimensionality before being entered into the fusion module. This step ensures balanced and effective multimodal integration.

\subsubsection{Swin Feature Extraction}
Compared to traditional CNN-based methods commonly used in this field, the Swin Transformer demonstrates superior capabilities in visual feature extraction. By employing a hierarchical architecture with shifted window attention mechanisms, the Swin Transformer captures both local and global dependencies efficiently. This design allows it to retain the fine-grained details of local features while incorporating contextual information across larger regions, surpassing CNNs that rely on fixed local receptive fields and hierarchical feature aggregation. This enables the Swin Transformer to extract richer and more robust features, providing a solid foundation for subsequent multimodal fusion tasks.

In this module, we use a 6-layer Swin Transformer as the image encoder, initialized with weights fine-tuned on emotion recognition dataset. The Swin Transformer serves as the backbone for extracting visual features, generating M feature tokens, each with dimension D, from the input facial images. These feature tokens retain hierarchical and contextual information, making them highly suitable for subsequent processing in the fusion stage.

\subsubsection{Multiple Instance Learning}
In the classical supervised 2-class classification problem, the objective is to develop a model that predicts a target variable $y \in {0, 1}$ for a given instance, $\mathbf{x} \in \mathbb{R}^D$. However, in the context of the MIL problem, the model is presented with a bag of instances $X = \{\mathbf{x}_1, \mathbf{x}_2, \dots, \mathbf{x}_q\}$, where the individual instances are independent of each other. The bag is associated with a single label $Y$ belonging to one of the two classes, i.e., $y \in {0, 1}$. Furthermore, although each instance within the bag is assumed to have its own label $y_1, y_2, \dots, y_q$, these labels are partially indicative of the bag’s overall label $Y$ but cannot be equated with it.

In MIL research, there is often a focus on understanding the relationships between instances within a bag and selecting a method that represents the bag in a more balanced or universal manner. The most common approach for this is pooling, such as average pooling or max pooling. With the advent of attention mechanisms, attention-based learnable pooling methods have emerged, one of the most notable being AMIL\cite{amil}. AMIL presents a simple yet effective approach that significantly enhances accuracy. It proposes an attention-based weighting method that calculates weights $a_1, a_2, \dots, a_q$ for the instances $\mathbf{x}_1, \mathbf{x}_2, \dots, \mathbf{x}_q$, then performs a weighted sum of the instances using these weights to obtain the pooled result for subsequent classification.

This study is primarily based on the AMIL approach, but with a key difference. While AMIL directly performs the weighted sum and classification after obtaining the weights, our method, which involves subsequent modality fusion with EEG data, avoids pooling into a single dimension at this stage. Therefore, instead of directly pooling, we select the top-K embeddings with the highest \( a_q \) values that best represent the subject's emotional state, which are then used for fusion. \( K \) is a hyperparameter representing the number of selected images. The formula is as follows:
\begin{equation}
\mathbf{z} = \left\{ \mathbf{x}_q \mid q \in \mathcal{T} \right\},
\end{equation}
where \( \mathcal{T} \) is the set of indices corresponding to the top \( K \) largest \( a_q \), defined as:
\begin{equation}
\mathcal{T} = \left\{ q \mid a_q \text{ is among the top } K \text{ largest values of } a_1, a_2, \dots, a_q \right\}.
\end{equation}
and the value of \( a_q \) is computed as:
\begin{equation}
a_q = \frac{\exp\big(\mathbf{w}^\top \tanh (\mathbf{V} \mathbf{x}_q^\top)\big)}{\sum_{j=1}^{q} \exp\big(\mathbf{w}^\top \tanh (\mathbf{V} \mathbf{x}_j^\top)\big)},
\end{equation}
where \( \mathbf{w} \in \mathbb{R}^{L \times 1} \) and \( \mathbf{V} \in \mathbb{R}^{L \times M} \) are parameters. This process allows the selected top-K images to effectively represent the entire bag.

\subsubsection{Cross Attention}
\begin{figure}[h]
\centering
\includegraphics[width=\linewidth]{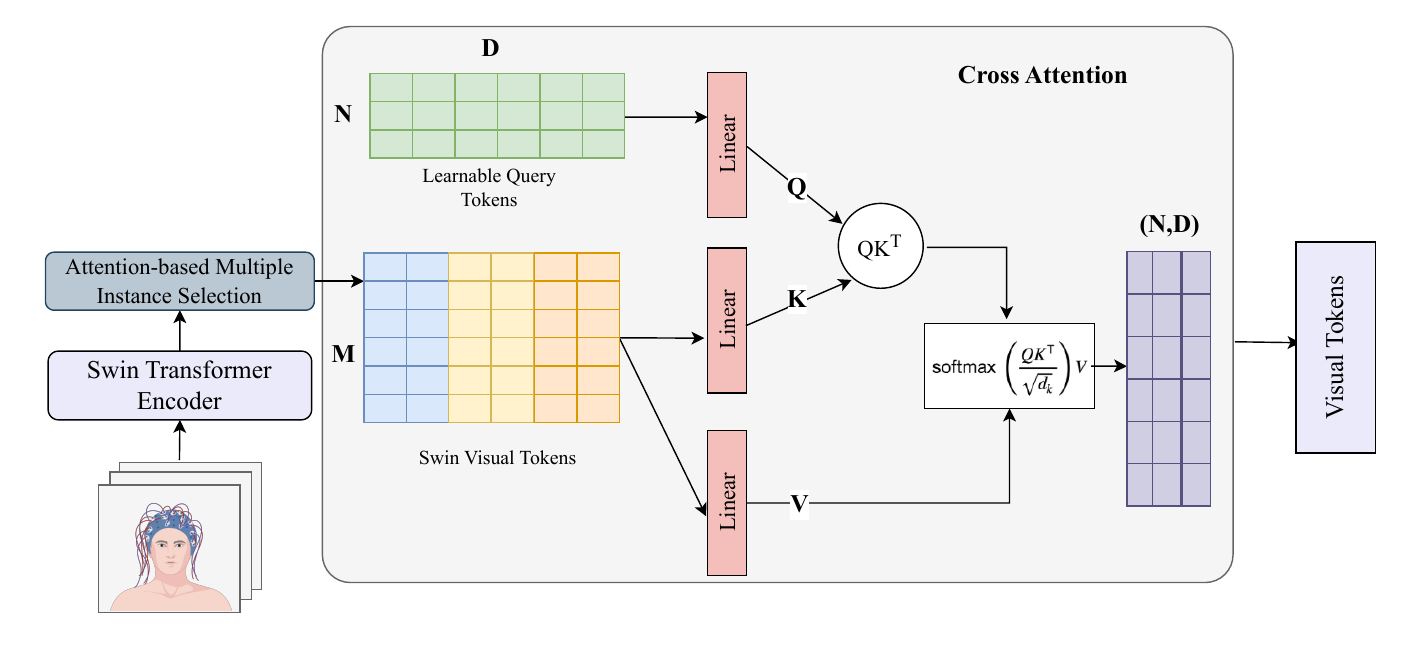}
\caption{The image feature dimensionality reduction module in this work. A cross-attention mechanism is used to reduce the number of feature vectors output by Swin Transformer to the predefined number of learnable query tokens. Here, M represents the number of feature vectors output by Swin Transformer, D is the vector dimension, and N is the number of learnable query tokens.}\label{fig2}
\end{figure}

In multi-modal fusion, the number of tokens provided by each modality plays a critical role in determining the quality of the fusion process. If the facial modality provides an excessively large number of tokens compared to the EEG modality, the transformer used for fusion may focus disproportionately on the visual features while neglecting those from the EEG signals. This imbalance can hinder the overall classification performance.

To address this, the cross-attention mechanism is employed to adjust the dimensionality of the feature vectors output by Swin. Specifically, the cross-attention mechanism compresses the M visual tokens into N learnable query tokens, where N $\ll$ M. This adjustment enables the visual tokens to match the token count of the EEG modality, achieving a more balanced and harmonious representation for fusion. This process ultimately contributes to improved classification performance by ensuring that both modalities are equally represented in the downstream tasks. As illustrated in Fig.~\ref{fig2}, the cross-attention mechanism shows as follows:

The learnable query tokens (N, D) and the visual tokens (M, D) extracted from the Swin transformer are first passed through linear layers to generate the query \(\mathbf{Q}\), key \(\mathbf{K}\), and value \(\mathbf{V}\) matrices, which are expressed as:
\[
\mathbf{Q} = \mathbf{W}_Q \mathbf{X}_q, \quad \mathbf{K} = \mathbf{W}_K \mathbf{X}_v, \quad \mathbf{V} = \mathbf{W}_V \mathbf{X}_v
\]
where \( \mathbf{X}_q \) represents the learnable query tokens of size (N, D), and \( \mathbf{X}_v \) represents the visual tokens extracted from the Swin transformer of size (M, D), with \( \mathbf{W}_Q \), \( \mathbf{W}_K \), and \( \mathbf{W}_V \) as the corresponding learnable weight matrices.

The attention output is then computed via the scaled dot-product attention mechanism:
\[
\mathbf{A} = \text{softmax} \left( \frac{\mathbf{Q} \mathbf{K}^T}{\sqrt{d_k}} \right) \mathbf{V}
\]
where \( \mathbf{A} \) is the reduced feature representation of size (N, D). 
This dimensionality reduction condenses the most relevant spatial and contextual information into a set of visual tokens, which helps balance the representation of the EEG and facial expression modalities, ensuring more harmonious multimodal fusion in subsequent tasks while reducing computational overhead.

To summarize, the facial feature extraction and balancing module utilizes the Swin Transformer for global feature representation, employs an AMIL-based method to select the most representative Top-K features, and incorporates a cross-attention mechanism to align and compress the visual tokens. This approach ensures a balanced token representation between the visual and EEG modalities, thereby enhancing the quality of multimodal fusion and improving overall classification performance.

\subsection{Fusion Module\label{sec:section34}}
In the proposed framework, the Fusion Module is designed to effectively combine information from EEG signals and facial features while preserving modality-specific characteristics. To achieve this, we use modal type embedding, position embedding, and a classification token (CLS token).

Modal Type Embedding: To differentiate between the two modalities, we introduce modal type embeddings. These are learnable vectors unique to each modality, appended to their respective input features. Modal-type embeddings provide the model with explicit information about the origin of each input token, enabling the Transformer to process and fuse modality-specific features more effectively. For instance, all EEG tokens are associated with a dedicated embedding vector, while facial tokens are assigned a distinct embedding, ensuring clear separation in the shared representation space.

Position Embedding: Position embeddings are applied to retain the order and structure of the input tokens for both EEG and facial features. EEG position embeddings encode temporal dependencies by representing the time-step information of each token, while facial position embeddings capture the spatial relationships among visual tokens extracted by Swin, preserving global contextual information essential for emotion recognition.

CLS Token for Classification: A CLS token is appended to the input sequence, serving as a global representation of the fused features across modalities. During the Transformer encoding process, the CLS token aggregates information from both EEG and facial feature tokens through attention mechanisms. In the output stage, the CLS token is passed to a fully connected layer for emotion classification. This approach allows the model to condense multimodal information into a compact and informative representation, optimizing the subsequent classification task.

Finally, the fused features are passed through a fully connected layer to produce the classification results. The cross-entropy loss is employed as the objective function to train the label classifier, and it is defined as follows: 
\begin{equation}
    Loss=-\frac{1}{batchsize} \sum_{j=1}^{batchsize}\sum_{i=1}^{n}y_{ji}log(\hat{y_{ji}}),    
\end{equation}
where $n$ is the number of classes; $y_{ji}$ and $\hat{y_{ji}}$ denote the true and predicted labels in a batch, respectively.

\section{Experiments and Result\label{sec:section4}}
In this section, we present the experimental results. First, we provide an overview of the DEAP dataset and describe the preprocessing methods applied to the data. Subsequently, we conduct extensive experiments on the DEAP dataset to evaluate and compare the classification performance of various multimodal fusion strategies. Furthermore, we investigate the impact of the token quantities from the two modalities on classification performance during multimodal fusion. Finally, we perform ablation studies to demonstrate the significance of each component in our proposed network architecture.

Thanks to the fast convergence of Milmer, the experimental trends and results become apparent within just 100 epochs. Loshchilov et al.\cite{loshchilov2016sgdr} introduced the cosine learning rate decay strategy, which has been widely validated for its effectiveness. All experiments in this section are conducted using the cosine learning rate decay strategy, with results reported over 100 epochs.

\subsection{Dataset and Settings\label{sec:section41}}
We employ the widely recognized DEAP dataset, which is commonly used in the field of multimodal affective computing, to evaluate the effectiveness of our proposed architecture. In the DEAP dataset, participants' emotions are elicited by watching music videos in a controlled laboratory environment, while their facial videos and bio-sensing signals are recorded synchronously. Following the viewing session, participants rate their emotional responses based on four dimensions: valence, arousal, dominance, and liking. Most studies in this area primarily focus on valence and arousal for emotion evaluation, and our study adheres to this convention.

The DEAP dataset provides continuous emotional ratings on a scale from 1 to 9, which are typically grouped into two, three, or four categories. For studies that focus on two- or three-class classification, valence and arousal are usually treated as independent dimensions. In the two-class approach, both dimensions are divided into high and low levels, while the three-class approach includes an additional neutral category. In contrast, the four-class classification treats valence and arousal as interrelated dimensions, dividing each of them into high and low levels, which results in a total of four distinct categories.

These three classification schemes—two-class, three-class, and four-class—are all commonly employed in the field. To enable comparison with prior research, we conduct experiments using all three classification schemes, referred to as DEAP-2, DEAP-3, and DEAP-4. The four-class scheme, being the most complex, captures the most nuanced emotional distinctions and is better suited for demonstrating differences in experimental results. In contrast, the two-class approach, though widely used, simplifies the task but may lead to overfitting and less persuasive results due to its reduced complexity. Therefore, the main results presented in this paper focus on the DEAP-4 experiment, while DEAP-2 and DEAP-3 are used primarily for comparison with other studies.

In the DEAP-4 setting, we use 5 as a threshold to classify emotions into four quadrants of the valence-arousal space: High Arousal High Valence (HAHV), High Arousal Low Valence (HALV), Low Arousal High Valence (LAHV), and Low Arousal Low Valence (LALV). The classification rule is defined as follows:
\begin{equation}
\text{Emotion} =
\begin{cases} 
\text{HAHV,} & a \geq 5 ,\ v \geq 5 \\ 
\text{HALV,} & a \geq 5 ,\ v < 5 \\ 
\text{LAHV,} & a < 5 ,\ v \geq 5 \\ 
\text{LALV,} & a < 5 ,\ v < 5
\end{cases}
\end{equation}
Here, a and v represent the arousal and valence scores, respectively.

For the DEAP-3 setting, we classify the emotional ratings into three categories: low (ratings from 1 to 3), medium (ratings from 4 to 6), and high (ratings from 7 to 9). In the DEAP-2 setting, the classification is simplified into two categories: low (ratings below 5) and high (ratings above 5).

The DEAP dataset consists of data from 32 participants, among whom only 22 provided facial image information. Each participant completed 40 trials. Notably, data from Participant 11 was excluded due to the absence of three facial expression videos, a decision made to prevent potential misalignment issues within the dataset.

In this study, the data is segmented into 3-second time windows, with EEG signals from each segment paired with 10 evenly spaced facial frames extracted over the 3-second duration. Google's Mediapipe library, built upon the efficient BlazeFace framework\cite{blazeface}, is utilized for face detection. The detected faces are resized to 224×224 pixels using center padding, with black borders added as needed to maintain aspect ratio. For the 32-channel EEG signals, preprocessing includes a bandpass filter within the 1–50 Hz range, followed by artifact removal using Fast ICA to eliminate noise caused by ocular and muscular activities, ensuring cleaner signals for subsequent analysis. Since the DEAP dataset has an EEG sampling rate of 128 Hz, resulting in 384 data points over a 3-second period, we employ a multi-layer perceptron (MLP) to upsample the signal length to 768, ensuring alignment with the image data.

\subsection{Comparison Experiments\label{sec:section42}}
The classification results of our proposed method, along with other studies that adopt the four-class approach, are presented in Table~\ref{table_deap4}. The results of studies employing the three-class approach are summarized in Table~\ref{table_deap3}, while the binary classification results for valence and arousal are shown in Table~\ref{table_deap2}.

\begin{table}[!htpb]
\caption{Comparison of emotion recognition performance across different studies on the DEAP dataset(DEAP-4 classification). This table presents various methods categorized by modality, including EEG-only and multimodal approaches, the number of emotion classes classified, and their corresponding accuracy and F1 scores. The best results are highlighted in \textbf{bold}, and differences from the best result are marked with downward arrows.}\label{table_deap4}
\centering
\resizebox{\textwidth}{!}{
\begin{tabular}{l c c c c}
\toprule
Method & Modality & Class & Accuracy(\%) & F1(\%) \\
\hline
Gupta wt al.(2018)\cite{gupta2018cross} & EEG & 4 
& $71.43_{\downarrow 25.29}$ & - \\
Kwon et al.(2018)\cite{kwon2018eeg} & EEG+GSR & 4
& $73.43_{\downarrow 23.29}$ & - \\
Marjit et al.(2021)\cite{marjit2021eeg} & EEG & 4 
& $81.25_{\downarrow 15.47}$ & $78.56_{\downarrow 18.15}$ \\
\midrule

Cimtay et al.(2020)\cite{Cimtay} & EEG+GSR+Face & 4 
& $53.87_{\downarrow 42.85}$ & - \\
Jung et al.(2019)\cite{jung2019utilizing} & EEG+Face & 4 
& $54.22_{\downarrow 42.50}$ & $31.00_{\downarrow 65.71}$ \\
Lee et al.(2024)\cite{lee2024emotion} & EEG+Face & 4 
& $83.20_{\downarrow 13.52}$ & $84.10_{\downarrow 12.62}$ \\
\midrule

Milmer (ours) & Multimodal & 4 & \textbf{96.72} & \textbf{96.71} \\
\bottomrule
\end{tabular}
}
\end{table}

The table presents various methods categorized by modality, including EEG-only and multimodal approaches, the number of emotion classes classified, and their corresponding accuracy and F1 scores. This comparison provides insights into the effectiveness of different methodologies under the DEAP-4 classification scheme.

Table~\ref{table_deap4} first presents three highly cited studies that focus exclusively on the EEG modality, without incorporating facial expressions, for emotion recognition. Among these studies, the highest performance achieved is an accuracy of 81.25\% and an F1 score of 78.56\%.

Following this, we compare several multimodal emotion recognition studies, primarily focusing on frameworks that integrate EEG and facial features. Notably, Cimtay et al. and Jung et al. are two highly cited works in this category, achieving accuracies of 53.87\% and 54.22\%, respectively, in the four-class classification task, which are lower compared to other studies. The underperformance of these models can be attributed to their reliance on conventional CNN-based feature extraction methods for each modality, which may not effectively capture the complex characteristics of EEG and facial expressions. Additionally, Cimtay et al. employ a decision tree mechanism for feature fusion, which struggles with high-dimensional and complex data, leading to suboptimal performance. 

A very recent study by Lee et al. presents a multimodal emotion recognition framework that effectively leverages contrastive learning and cross-modal attention mechanisms to enhance inter-modal feature alignment. Their use of contrastive learning stands out as a significant strength, enabling the model to learn richer and more discriminative multimodal representations. However, their approach to visual feature extraction is relatively simplistic, relying solely on a ViT-based encoder without further refinement, which may limit the potential of their model. Consequently, their framework achieves an accuracy of 83.20\% and an F1 score of 84.10\%.

Finally, our proposed framework, Milmer, achieves the best results with an accuracy of 96.72\% and an F1 score of 96.71\%, demonstrating the superiority of our approach. These results validate the effectiveness of our feature extraction and fusion strategies in improving emotion recognition performance.

\begin{table}[!htpb]
\caption{Comparison of emotion recognition performance across different studies on the DEAP dataset (DEAP-3 classification). The best results are highlighted in \textbf{bold}, with performance differences indicated by downward arrows.}\label{table_deap3}
\centering
\resizebox{\textwidth}{!}{
\begin{tabular}{l c c c c}
\toprule
Method & Modality & Class & Accuracy(\%) & F1(\%) \\
\hline
Wu et al.(2023)\cite{wu2023recognizing} & EEG+Face & 3 
& $71.48_{\downarrow 25.36}$ & - \\
Lee et al.(2024)\cite{lee2024emotion} & EEG+Face & 3 
& $88.95_{\downarrow 7.89}$ & $88.90_{\downarrow 7.99}$ \\
Husformer(2024)\cite{husformer} & EEG+EMG+EOG+GSR & 3 
& $91.00_{\downarrow 5.84}$ & $91.05_{\downarrow 5.84}$ \\
Muhammad et al.(2023)\cite{muhammad2023bimodal} & EEG+Face & 3 
& $91.54_{\downarrow 5.30}$ & $91.07_{\downarrow 5.82}$ \\
\midrule

Milmer (ours) & Multimodal & 3 & \textbf{96.84} & \textbf{96.89} \\
\bottomrule
\end{tabular}
}
\end{table}

Table~\ref{table_deap3} presents the experimental results of various multimodal learning studies conducted on the DEAP-3 classification scheme. Wu et al. proposed an innovative approach by leveraging Action Units (AUs) to represent facial expressions and designing a multimodal learning framework based on CNN and LSTM, achieving an accuracy of 71.48\%. However, their feature fusion strategy remains relatively simplistic, and the use of AUs alone may not fully capture the richness of facial expression information, potentially limiting the model’s overall performance.

Two recent studies, Husformer and Muhammad et al., have demonstrated significant advancements, achieving accuracies exceeding 90\%, thereby outperforming previous works. Husformer employs a cross-attention-based fusion strategy to effectively integrate multimodal information, while Muhammad et al. utilize DeepCCA for feature fusion. These advanced fusion techniques contribute to their superior performance. However, Husformer does not incorporate facial expression features, which are crucial emotional cues; thus, relying solely on physiological signals may restrict its ability to comprehensively capture the emotional states of subjects. On the other hand, Muhammad et al.’s approach, despite its strong fusion capabilities, faces challenges in feature extraction from individual modalities, which could limit further performance improvements.

Finally, our proposed framework, Milmer, achieves the highest performance on the DEAP-3 dataset, with an accuracy of 96.84\% and an F1 score of 96.89\%, demonstrating its superiority over existing approaches.

\begin{table}[!htpb]
\caption{Comparison of binary classification performance across different studies on the DEAP dataset (DEAP-2 classification). This table presents methods categorized by modality, the number of emotion classes classified, and their corresponding accuracy for arousal and valence classification. The best results are highlighted in \textbf{bold}, with performance differences indicated by downward arrows.}\label{table_deap2}
\centering
\resizebox{\textwidth}{!}{
\begin{tabular}{l c c c c}
\toprule
Method & Modality & Class & Arousal(\%) & Valence(\%) \\
\hline

Romeo et al.(2019)\cite{romeo2019multiple} & EEG & 2 
& $61.10_{\downarrow 36.81}$ & $54.60_{\downarrow 43.24}$ \\
Salama et al.(2018)\cite{salama2018eeg} & EEG & 2 
& $87.97_{\downarrow 9.94}$ & $86.00_{\downarrow 11.84}$ \\
\midrule

Wu et al.(2023)\cite{wu2023recognizing} & EEG+Face & 2 
& $72.73_{\downarrow 25.18}$ & $64.77_{\downarrow 33.07}$ \\
Huang et al.(2019)\cite{huangEnhance} & EEG+Face & 2 
& $74.23_{\downarrow 23.68}$ & $80.30_{\downarrow 17.54}$ \\
Jung et al.(2019)\cite{jung2019utilizing} & EEG+Face & 2 
& $78.34_{\downarrow 19.57}$ & $79.52_{\downarrow 18.32}$ \\
Wang et al.(2025)\cite{wang2025design} & EEG+Face & 2 
& $83.24_{\downarrow 14.67}$ & $86.75_{\downarrow 11.09}$ \\
Zhao et al.(2021)\cite{zhao2021expression} & EEG+Face & 2 
& $86.80_{\downarrow 11.11}$ & $86.20_{\downarrow 11.64}$ \\
Wu and Li(2023)\cite{wu2023multi} & EEG+Face & 2 
& $94.94_{\downarrow 2.97}$ & $95.30_{\downarrow 2.54}$ \\
Wang et al.(2023)\cite{wang} & EEG+Face & 2 
& $97.15_{\downarrow 0.76}$ & $96.63_{\downarrow 1.21}$ \\
Hosseini et al.(2024)\cite{greatHelp} & EEG+Face & 2 
& $97.79_{\downarrow 0.12}$ & $97.39_{\downarrow 0.45}$ \\
\midrule

Milmer (ours) & Multimodal & 2 & \textbf{97.91} & \textbf{97.84} \\
\bottomrule
\end{tabular}
}
\end{table}

To facilitate comparisons with the most studies in the field, we conducted experiments on the DEAP-2 dataset, as shown in Table~\ref{table_deap2}. The overall accuracy of the two-class task is generally higher than that of the previously discussed three and four-class tasks. Among the listed studies, Wu et al. and Jung et al., which have been previously discussed, employ multiple classification strategies and will not be elaborated further here.

Huang et al. adopt a CNN-based feature extraction approach and utilize a weighted summation strategy for modality fusion, achieving arousal and valence accuracies of 74.23\% and 80.30\%, respectively. In contrast, Wang et al., Zhao et al., and Wu and Li employ LSTM-based feature fusion methods. Notably, Zhao et al. enhance the performance by incorporating an attention mechanism within the LSTM framework, resulting in improved classification accuracy.

Finally, Wang et al., Hosseini et al., and our proposed Milmer achieve very similar results, making it challenging to directly compare their advantages and limitations. This observation highlights one of the challenges associated with the two-class task, where performance differences become less distinguishable. Wang et al. employ a CNN-based feature extraction method for both EEG and facial features but enhance the architecture with attention mechanisms and a more complex CNN structure. Hosseini et al. introduce a novel loss function and propose a qualitative and quantitative evaluation of emotions, demonstrating an innovative approach.

Compared to these methods, our proposed framework, Milmer, offers advantages through the incorporation of multi-instance learning (MIL), Swin Transformer-based feature extraction, and cross-attention-based modality balancing, ultimately achieving the highest arousal and valence accuracies of 97.91\% and 97.84\%, respectively.

\begin{figure}[t]
\centering
\includegraphics[width=\linewidth]{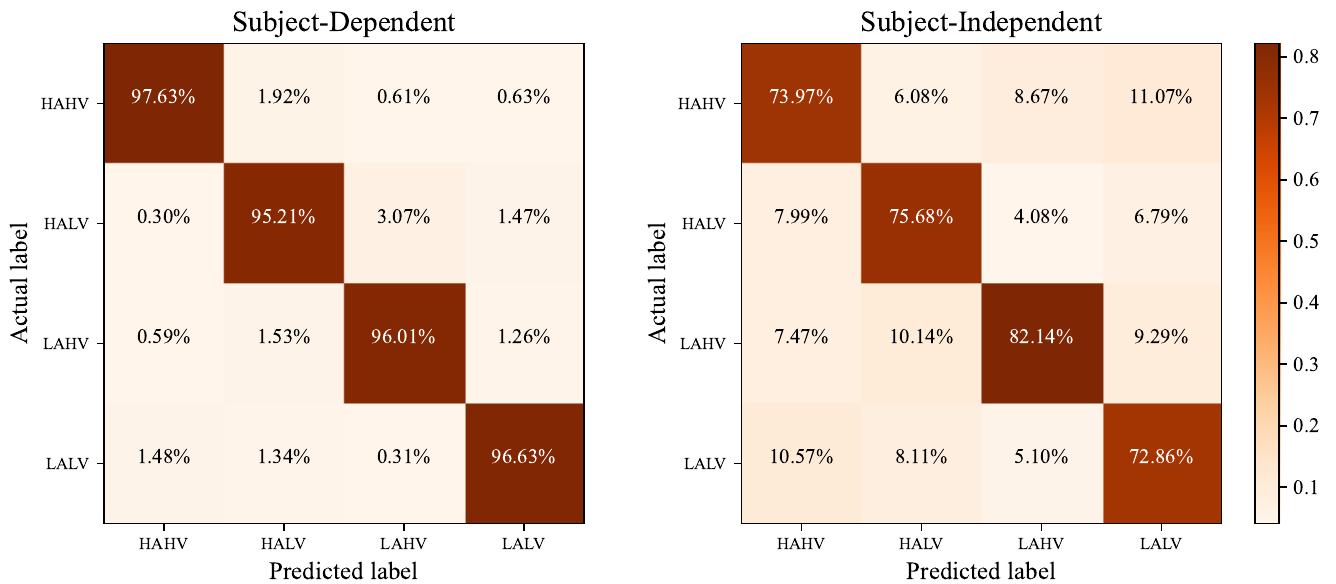}
\caption{Confusion matrices of subject-dependent and subject-independent experiments.}\label{fig3}
\end{figure}

Furthermore, confusion matrices are presented for the experiments conduct under subject-dependent and subject-independent settings, as depicted in Fig.~\ref{fig3}. The results demonstrate that our model performs exceptionally well in the subject-dependent setting, achieving high recognition accuracy across all emotion categories. This highlights the effectiveness of the proposed approach in capturing intra-subject patterns in EEG signals and facial expressions.

In subject-independent experiments, although the model demonstrates notable improvements compared to previous studies, a significant performance gap persists when compared to the subject-dependent setting. This discrepancy can be attributed to the substantial individual differences in both EEG signals and facial expressions, which pose a considerable challenge to the model's generalization capability. This issue is a common challenge faced in the field of multimodal emotion recognition research.

In general, these results underscore the strengths of the proposed method in controlled conditions while highlighting the inherent difficulties of cross-subject emotion recognition tasks. Future work could focus on strategies to further enhance the model's robustness to individual variability.

\subsection{Visual Feature Integration Analysis\label{sec:section43}}

\begin{table}[!htpb]
\caption{This table compares three strategies for integrating visual features extracted from multiple images before fusing them with EEG features in a cross-modal framework: (1) no integration (None), (2) two fully connected layer (MLP), and (3) cross-attention (CA). The table also examines the effect of varying the output size of the integrated features on the final classification performance. The best results are highlighted in \textbf{bold}, and differences from the best result are marked with downward arrows.}\label{table_CA}
\centering
\begin{tabular}{c c c c}
\toprule
Fusion Methods & Output Size & Accuracy(\%) & F1(\%) \\
\hline
None & 147 & $89.91_{\downarrow 6.81}$ & $89.35_{\downarrow 7.36}$\\
MLP & 147 & $88.83_{\downarrow 7.89}$ & $89.16_{\downarrow 7.55}$\\
\midrule
CA & 196 & $94.25_{\downarrow 2.47}$ & $94.27_{\downarrow 2.44}$ \\
CA & 147 & $95.51_{\downarrow 1.21}$ & $95.87_{\downarrow 0.84}$ \\
CA & 128 & $95.99_{\downarrow 0.73}$ & $95.92_{\downarrow 0.79}$ \\
CA & 96 & $96.23_{\downarrow 0.49}$ & $96.42_{\downarrow 0.29}$ \\
CA (Milmer) & 64 & \textbf{96.72} & \textbf{96.71} \\
CA & 32 & $95.05_{\downarrow 1.67}$ & $95.25_{\downarrow 1.46}$\\
CA & 16 & $93.84_{\downarrow 2.88}$ & $93.99_{\downarrow 2.72}$\\
\bottomrule
\end{tabular}
\end{table}

Table~\ref{table_CA} presents a comparison of three strategies for integrating visual features extracted from multiple facial images before fusing them with EEG features in a cross-modal framework: (1) no integration, (2) MLP, and (3) cross-attention (CA). In the no integration approach, visual features are directly passed to the next stage of multimodal fusion without additional processing. In contrast, the MLP approach linearly maps the visual features to a four-fold higher dimension before reducing them back. The results demonstrate that employing cross-attention significantly improves classification performance, far outperforming both no integration and MLP, thereby highlighting the importance of CA in this framework.

Furthermore, the table investigates the impact of varying the number of facial modality tokens on classification performance. In addition to the original length of 147, six additional configurations were evaluated. The results reveal that excessively increasing the dimension of facial features leads to degradation of performance. This aligns with our hypothesis that an overly high-dimensional visual representation may dominate the transformer's attention mechanism, potentially overshadowing the contribution of EEG features. On the other hand, when the number of visual tokens is reduced to within three times the EEG token count, promising classification results are achieved. However, further reducing the number of visual tokens below that of EEG results in decreased performance. Ultimately, the best performance, with an accuracy of 96.72\%, is achieved when the visual token output size is set to 64. This suggests that a balanced configuration allows the model to effectively leverage both modalities without introducing significant bias toward one.

These findings emphasize the importance of careful tuning of token dimensions in cross-modal fusion tasks to ensure that critical information from both modalities is preserved and utilized effectively.

\subsection{Ablation Study\label{sec:section44}}

\begin{table}[!htpb]
\caption{Ablation Study on the Contribution of EEG Data, Facial Data, and Fusion Methods in Model Performance. The best results are highlighted in \textbf{bold}, and differences from the best result are marked with downward arrows.}\label{table_ab1}
\centering
\begin{tabular}{c c c c c}
\toprule
Facial Data & EEG Data & Fusion Methods & Accuracy(\%) & F1(\%) \\
\midrule
× & \checkmark & - 
& $62.33_{\downarrow 34.39}$ & $60.93_{\downarrow 36.32}$ \\
\checkmark & × & - 
& $83.59_{\downarrow 13.13}$ & $83.67_{\downarrow 13.04}$ \\
\hline

\checkmark & \checkmark & DeepCCA 
& $88.13_{\downarrow 8.59}$ & $88.74_{\downarrow 7.97}$ \\
\checkmark & \checkmark & Concatenation 
& $89.88_{\downarrow 6.84}$ & $90.28_{\downarrow 6.43}$ \\
\checkmark & \checkmark & Transformer (Milmer) 
& \textbf{96.72} & \textbf{96.71} \\

\toprule[1.0pt]
\end{tabular}
\end{table}

To validate the contributions of the three critical components in our framework, the EEG module, the facial feature extraction and balancing module, and the modality fusion module, we conducted ablation studies, as shown in Table~\ref{table_ab1}. We assessed the classification performance of each modality independently and compared our transformer-based fusion approach with two commonly used fusion methods in the field.

The results presented in Table~\ref{table_ab1} indicate that when only the EEG modality is used, the framework achieves an accuracy of 62.33\% and an F1 score of 60.93\%, while the facial modality alone yields slightly better performance, with an accuracy of 83.59\% and an F1 score of 83.67\%. The relatively lower performance of the EEG-only model can be attributed to the simplified EEG processing adopted in our study, as our primary focus lies in cross-modal fusion rather than sophisticated EEG signal processing.

When both modalities are incorporated, performance improves significantly, with accuracy exceeding 88\% across different fusion methods. Specifically, the DeepCCA and Concatenation methods achieve accuracies of 88.13\% and 89.88\%, respectively, which are notably lower compared to the 96.72\% accuracy and 96.71\% F1 score achieved by our transformer-based fusion approach.

These findings highlight the complementary nature of EEG and facial features and emphasize the crucial role of the fusion module in effectively harnessing their synergy. The superior performance of the transformer-based approach underscores its potential in cross-modal fusion tasks, demonstrating its ability to capture complex interactions between modalities more effectively than traditional methods.

\begin{table}[!htpb]
\footnotesize
\centering
\caption{Ablation study on the effectiveness of the facial feature extraction and balancing module, which consists of three submodules: fine-tuned (FT) Swin feature extraction, multiple instance learning (MIL), and cross-attention (CA). The table compares different combinations of these submodules to assess their individual and joint contributions to classification performance. The best results are highlighted in \textbf{bold}, and differences from the best result are marked with downward arrows.}\label{table_ab2}
\begin{tabular}{c c c c c c c}
\toprule[1.0pt]
\multicolumn{2}{c}{{Methods}} & \multicolumn{1}{c}{FT}  & \multicolumn{1}{c}{MIL}&\multicolumn{1}{c}{CA}& \multicolumn{1}{c}{Accuracy(\%)} & \multicolumn{1}{c}{F1(\%)} \\
\cline{1-7}

\hline
\multicolumn{2}{c}{Single Picture} &× & × & ×	
& $86.39_{\downarrow 10.33}$ & $86.09_{\downarrow 10.62}$ \\ 
\multicolumn{2}{c}{Multiple Picture} &× & × & ×	
& $86.55_{\downarrow 10.17}$ & $85.94_{\downarrow 10.77}$ \\ 

\hline
\multicolumn{2}{c}{FT}  & \checkmark & × & × 
& $88.42_{\downarrow 8.30}$ & $88.57_{\downarrow 8.14}$ \\ 
\multicolumn{2}{c}{MIL} & × & \checkmark & × 
& $89.83_{\downarrow 6.89}$ & $90.29_{\downarrow 6.43}$ \\ 
\multicolumn{2}{c}{CA}  & × & × & \checkmark
& $87.64_{\downarrow 9.08}$ & $87.78_{\downarrow 8.93}$ \\ 

\hline
\multicolumn{2}{c}{FT + MIL} & \checkmark & \checkmark	& ×	
& $89.91_{\downarrow 6.81}$ & $89.35_{\downarrow 7.36}$ \\ 
\multicolumn{2}{c}{FT + CA}  & \checkmark & × & \checkmark	
& $90.01_{\downarrow 6.71}$ & $90.06_{\downarrow 6.65}$ \\ 
\multicolumn{2}{c}{MIL + CA} & × & \checkmark & \checkmark	
& $94.46_{\downarrow 2.26}$ & $94.58_{\downarrow 2.13}$ \\ 

\hline
\multicolumn{2}{c}{Milmer (ours) } & \checkmark	& \checkmark & \checkmark	
& \textbf{96.72} & \textbf{96.71}\\ 

\toprule[1.0pt]
\end{tabular}
\label{table6}
\end{table}

Furthermore, our facial feature extraction and balancing module is composed of three key submodules: fine-tuned (FT) Swin feature extraction, multiple instance learning (MIL), and cross-attention (CA). To validate the contributions of these components, we conducted ablation studies, as presented in Table~\ref{table_ab2}.

The results demonstrate that whether using the conventional single-image representation of facial features or our multi-image approach, the absence of the facial feature extraction and balancing module results in an accuracy of approximately 86\%. 

Individually adding the FT, MIL, and CA submodules results in accuracies of 88.42\%, 89.83\%, and 87.64\%, respectively, showing incremental improvements over the baseline. When FT is combined with MIL or CA, the accuracy increases to 89.91\% and 90.01\%, respectively, slightly exceeding the performance achieved using FT alone. However, the combination of CA and MIL leads to a significant performance boost, achieving an accuracy of 94.46\% and an F1 score of 94.58\%, substantially outperforming individual module contributions. This improvement can be attributed to the strong correlation between CA and MIL, as the primary role of the CA module is to effectively integrate the multi-instance outputs generated by the MIL module. 

Ultimately, the best results are obtained when all three submodules—FT, MIL, and CA—are incorporated, achieving an accuracy of 96.72\% and an F1 score of 96.71\%, demonstrating the effectiveness of our proposed module design. These findings underscore the importance of integrating multiple complementary strategies within the facial feature extraction and balancing module. The synergy among FT, MIL, and CA enables the model to leverage richer and more discriminative facial representations, contributing to the overall robustness and accuracy of the emotion recognition framework.

\section{Conclusion and Future work\label{sec:section5}}
This paper proposes a comprehensive method for assessing the credibility of Internet of Things devices. We evaluate multiple dimensions of device network behaviors and utilize a Transformer to capture the temporal variability of network features and to predict the features for the next time step. The predicted behavior information is then compared with the actual collected behavior information to calculate similarities and differences. These metrics are used to ascertain the credibility of the device. Finally, our method's reliability is validated through experimental testing, demonstrating its effectiveness in accurately assessing the trustworthiness of IoT devices.This study proposes a multimodal learning framework based on EEG and facial expressions, integrating feature extraction and deep learning for emotion recognition. For the EEG modality, widely adopted methods in the field were employed, while more advanced feature extraction techniques were introduced for facial expression processing. Additionally, a more sophisticated fusion module was utilized to explore effective strategies for integrating features from the two modalities. Extensive experiments conducted on the DEAP dataset demonstrate that the proposed framework surpasses state-of-the-art methods in terms of accuracy. Furthermore, ablation studies on individual modules validate the effectiveness of our approach.

This study addresses the challenges of emotion recognition by integrating facial expression analysis with EEG signals, introducing a novel framework—Milmer. To overcome the limitations observed in previous research, Milmer proposes several optimization strategies.

First, prior studies often employed overly simplistic and coarse multimodal fusion techniques, neglecting the substantial differences between EEG and facial expressions. In contrast, Milmer adopts a transformer-based fusion approach, utilizing cross-attention to balance token distribution during modality fusion, thereby enhancing the integration of these heterogeneous data sources (see Section~\ref{sec:section33} and Section~\ref{sec:section34}). Experimental results validate the effectiveness of our fusion and balancing strategies, as demonstrated in Section~\ref{sec:section43} and Section~\ref{sec:section44}.

Second, traditional facial expression feature extraction methods rely on outdated techniques and typically fail to incorporate temporal dynamics by using multiple images over time, thus missing critical sequential information. To address this gap, we propose a multi-instance learning based approach (see Section~\ref{sec:section33}), which is further validated through ablation studies, demonstrating its contribution to the overall performance (see Section~\ref{sec:section44}).

Furthermore, we conducted comprehensive comparisons with numerous well-established studies in the field, covering both unimodal and multimodal approaches across various classification tasks. Extensive experiments on the DEAP dataset confirm the superiority of our proposed framework, achieving an average classification accuracy of 96.72\% in the four-class emotion classification task, surpassing existing methods and further validating the effectiveness of our approach (see Section~\ref{sec:section42}).

For future work, integrating additional physiological signals, such as electrocardiogram (ECG) and galvanic skin response (GSR), could further enhance the diversity and robustness of multimodal learning frameworks. Continuous advancements in feature extraction techniques for individual modalities, highlight the need to stay abreast of developments and refine unimodal processing approaches. Moreover, the design of classification tasks and loss functions for emotion recognition presents opportunities for further optimization, offering valuable insights into the burgeoning field of multimodal learning. Finally, the joint recognition of facial expressions and physiological signals remains a challenging yet underexplored area, warranting deeper and more comprehensive research in the future.

\section{Acknowledgment\label{sec:section6}}
This work is supported by the Beijing Natural Science Foundation (no.4254091).

\bibliography{main}
\bibliographystyle{ieeetr}

\end{document}